\documentclass[letterpaper]{article}
\usepackage{proceed2e}
\usepackage[margin=1in]{geometry}
\usepackage{graphicx}
\usepackage{amsmath, amssymb}
\usepackage{twoopt}
\usepackage{subcaption}
\usepackage{float}
\usepackage{algorithm, algorithmic}
\usepackage{url}

\usepackage{times}

\title{Learning the Base Distribution in Implicit Generative Models}

\author{ { \thanks{This work is supported by NSF grant \#1319708.}}Y. Cem Subakan$^\flat$, Oluwasanmi Koyejo$^\flat$, Paris Smaragdis$^{\flat, \sharp}$ \\
        $^\flat$UIUC, $^\sharp$Adobe Inc.} % LEAVE BLANK FOR ORIGINAL SUBMISSION.
\begin{document}

\maketitle
\begin{abstract}
Popular generative model learning methods such as Generative Adversarial Networks (GANs), and Variational Autoencoders (VAE) enforce the latent representation to follow simple distributions such as isotropic Gaussian. In this paper, we argue that learning a complicated distribution over the latent space of an auto-encoder enables more accurate modeling of complicated data distributions. Based on this observation, we propose a two stage optimization procedure which maximizes an approximate implicit density model. We experimentally verify that our method outperforms GANs and VAE on two image datasets (MNIST, CELEB-A). We also show that our approach is amenable to learning generative model for sequential data, by learning to generate speech and music.
\end{abstract}

\section{Introduction}
Generative model learning is the task where the goal is to learn a model to generate artificial samples which follow the underlying probability density function of a given dataset. When the dataset comprises of scalars, or of low dimensional (2-3 dimensions) vectors and follow a unimodal distribution, one can use a simple density model such as the multivariate Gaussian, and fit the model to the data using maximum likelihood. Unfortunately, such simple densities do not have sufficient expressive power to learn the distribution of more complicated data such as natural images, or audio because of the aforementioned high dimensional and multi-modal nature of the data. 

There exists several generative model learning methods in the machine learning literature. One way of approaching the problem is to use a linear latent variable model (LVM) such as a mixture model \cite{Bishop2006}, a latent factor model such as probabilistic PCA \cite{Tipping1999}, Hidden Markov model (HMM) \cite{Rabiner1989}, or linear dynamical systems \cite{Bishop2006, Roweis99}. These models can successfully capture the multi-modality, or low rank nature of the datasets, however they rely on linear and tractable forward mappings, and therefore lack the expressive power of modern neural network models. 

More recently, the mainstream approaches for learning a generative model for complicated datasets have been centered around models that combine latent variable modeling with non-linear neural network mappings. One prominent example of such approaches is Variational Autoencoders (VAEs) \cite{Kingma2013}. VAEs consider a latent variable model where the latent representation is mapped to the observation space via a complicated neural network. The variational expectation maximization algorithm in \cite{Kingma2013} maximize a variational lower bound on the maximum likelihood objective. The prior distribution is typically chosen as a simple distribution such that the KL-divergence term in the lower bound is tractable.  %The model likelihood is approximated with a variational Expectation Maximization algorithm where a lower bound to the likelihood is iteratively maximized. Talk about maximumum likelihood here (similarly for GANs maybe say it is not maximum likelihood, then say that we do max likelihood)  
In this paper we argue that using a simple prior distribution is detrimental to the overall quality of the learned generative model. 

Another very popular method that also uses a restricted latent representation is Generative Adversarial Networks (GANs) \cite{Goodfellow2014}. The main conceptual differences of GANs from typical latent variable models (including VAEs) is that GANs are an implicit generative model learning methodology \cite{Mohamed2017}, where the model distribution is defined without specifying an output density. More importantly, unlike LVMs GANs do not maximize the standard maximum likelihood objective. Instead, GANs approximate the underlying dataset density via an additional discriminator network. Although an appealing idea, GANs are incredibly hard to train (as evidenced by the sheer number of GAN training papers in the last few years), and suffer from the predictable mode collapse problem (We delve more into this in the main text). 

In this paper, we propose an implicit generative model learning method which maximizes the maximum likelihood training objective. Unlike GANs, the method does not rely on auxiliary networks such as discriminator or critic networks. For training, we propose a simple two stage training method, which maximizes a maximum likelihood training objective, and therefore does not suffer from the mode collapse problem that GANs are notorious for. 

%The proposed algorithm is simple. We first fit 
\newcommand{\pdata}[1][x]{p_\text{data}(#1)} 
\newcommandtwoopt{\pmodel}[2][x][\theta]{p_\text{model}(#1|#2)}
\newcommand{\pfor}{p_\text{forward}(x|h, \theta)}
\newcommand{\pri}{p(h)}
\newcommand{\pout}{p_\text{out}}
\newcommandtwoopt{\pbase}[2][p^0][h]{#1(#2)}

\section{Generative Model Learning} 
The purpose of this section is to set the notation and the required concepts before we formally introduce our algorithm. As we discussed in the introduction, the goal in generative model learning is to approximate the underlying data density $\pdata$ with the density that our model implies, which we denote by $\pmodel$, where $\theta$ denotes the model parameters. Maximum likelihood training minimizes the Kullback-Leibler (KL) divergence between the data density and model density: 
\begin{align}
	&\min_\theta \; \text{KL}(\pdata \| \pmodel ) \notag \\
	=& \min_\theta \int \pdata \log \frac{\pdata}{\pmodel} dx \notag  \\
	\propto& \min_\theta - \int \pdata \log \pmodel dx \notag \\
	\approx& \max_\theta \sum_n \log p_\text{model}( x_n | \theta) \label{eq:mlobj}
\end{align}
where the last step is a Monte Carlo approximation to the integral, and we recognize Equation \eqref{eq:mlobj} as the maximum likelihood objective. Note that $x \in \mathbb R^L$ denotes the variable we use to denote the observation space, and we use the subscripted version $x_n$ to denote the data item with index $n$. 

It is usually not easy to compute (not tractable) the likelihood function $\pmodel$ unless we work with very simple models. In LVMs, Jensen's inequality is used to compute a lower bound to the maximum likelihood objective:
\begin{align} 
	&\log \pmodel  = \log \int \pfor \pri dh,  \notag \\
	&= \log \int \frac{\pfor \pri}{ q(h)} q(h) dh, \notag \\
	&\geq \mathbb E_{q(h)}[ \log \pfor ] - \text{KL}(q(h)\| \pri),  \label{eq:vlb} 
\end{align} 
where Equation \eqref{eq:vlb} is known as the variational lower bound \cite{Bishop2006}, or ELBO \cite{Hoffman2013}, where $h \in \mathbb R^{K}$ denotes the latent variable, and $q(h)$ denotes the variational distribution over the latent variable. In linear LVMs with tree structured latent variables (e.g. mixture models, HMMs), we can use the posterior $p(h|\theta)$ as the variational distribution, because the posterior makes this bound tight \cite{Murphy2012, Bishop2006}. 

In the general situation where the forward mapping is defined via a non-linear mapping, such that $\pfor = \pout (x;f_\theta(h))$, where $f_\theta(h): \mathbb R^K \to \mathbb R^L$ is the nonlinear deterministic mapping, and $\pout(.)$ is the employed noise model, computing the posterior distribution is not analytically tractable in general. VAEs therefore use a neural network mapping for the variational distribution $q_\phi(h) = \mathcal N(x; \mu_\phi (x), \sigma^2_\phi (x) I )$, where $\mathcal N(.)$ denotes the Normal distribution and the neural network mappings $\mu_\phi(x), \sigma^2_\phi(x) : \mathbb R^L \to \mathbb R^K$ parametrize the variational distribution.  

Although the likelihood computation in VAEs is intractable and require the variational EM algorithm described above, we argue in this paper that the main failure mode of VAEs is caused by the simplistic prior choices for $\pri$, as we demonstrate this in the experiments section. 

Another popular way to learn generative models is via GANs. GANs are implicit generative models, therefore they do not employ an output distribution $\pout$. Namely, the data generation mechanism is defined as follows: 
\begin{align}
	h \sim \pbase, \; x = f_\theta(h), \label{eq:implmodel}
\end{align}
where we call $\pbase$ \emph{the base distribution}, typically chosen as a simplistic distribution such as an isotropic Gaussian distribution, and $f_\theta(h)$ is a deterministic forward mapping similar to what we have denoted for VAEs above. GANs therefore do not employ an output distribution $\pout(.)$, but rather define $\pmodel$ via a deterministic transformation of the base distribution $\pbase$. 

In this paper, we also argue that one of the reasons why GANs might underperform is because of the simplistic base distribution choice. In addition to this, GANs also complicate the model parameter optimization by introducing a discriminator network. GANs in their original formulation \cite{Goodfellow2014}, approximate the ratio between the data density and the model density \cite{Mohamed2017}: 
\begin{align}
	\mathcal L(\theta, \xi) =& \sum_n \log D_\xi(x_n) + \sum_{n'} \log 1 - D_\xi(x'_{n'}) \notag \\
	\to & \sum_n \log \frac{\pdata[x_n]}{\pmodel[x_n] + \pdata[x_n] }  \notag \\
	& + \sum_{n'} \log \frac{\pmodel[x_{n'}']}{\pmodel[x_{n'}'] + \pdata[x_{n'}'] }
\end{align}
where $x_n$ denotes the training instances, and $x_n'$ denotes samples generated from the model. The convergence to the second line (which can be recognized as the Monte Carlo estimate for the Jensen-Shannon divergence) can be easily seen by maximizing the objective $\mathcal L(\theta, \xi)$ with respect to the discriminator parameters $\xi$ \cite{Goodfellow2014}. The big conceptual problem with GANs is that the optimization step for the generator parameters cause mode collapse. This can be easily seen by examining the corresponding loss function. The original paper suggests the maximization of the following objective: 
\begin{align*}
	&\max_\theta \sum_{n'} D(x'_{n'}), \; x'_{n'} \sim \pmodel \notag \\ 
  \approx &\max_\theta \int \pmodel \log \frac{\pdata}{\pdata + \pmodel}dx 
  % \leq &\max_\theta \int \pmodel \log \frac{\pdata}{ \pmodel}dx \\
  % = & \min_\theta KL( \pmodel \| \pdata ),
\end{align*}
where we assumed that the discriminator is trained until convergence. We can see that the objective in the last equation has a mode seeking/zero avoiding behavior, similar to $\text{KL}(\pmodel\| \pdata)$ \cite{minka2001}. In practice, therefore the discriminator is not trained until convergence, and there are various heuristics that tries to deal with mode collapse \cite{Salimans2016}. 

There exists several other variants of GANs which use other divergences \cite{Nowozin2016}, or which are based on approximate optimal transport metrics \cite{Arjovsky2017, salimans2018}. Or, some approaches use a GAN ensemble to approximate the whole density \cite{Saatci2017}. 

In this paper, we propose a much simpler approach, which optimizes a maximum likelihood objective using an implicit density model. The optimization does not involve an additional discriminator, and the approach does not suffer from mode collapse since it maximizes a maximum likelihood objective. 

We would also like to point out that there is a recent work on generative model learning, which does maximum likelihood for implicit models \cite{Dinh2016} for certain types of invertible mappings such as convolutions. However, they do not consider general mappings as we do. In addition to this we advocate using multi-modal distributions in the latent space in this paper.  

%or other tricks to make the optimization easier , or alleviate the mode collapse problem [gan training trick paper]. But there has not been enough focus on achieving a more expressive distribution via learning the base distribution. 

\section{Learning in Implicit Generative Models} 
%As we mentioned before GANs are only one option for learning implicit generative models. According to \cite{Mohamed2013}, there are three other broad categories which are ratio matching [ ], divergence matching [ ] and moment matching [ ]. The first two methods are basically variations on the GAN score function, and do not consider the effect of base distribution. Moment matching is a simple method where the observable moments of the model density are matched with the moments computed from data. In our experience this method's performance highly depends on the quality of the choice of the moments. In this paper, we propose a novel maximum likelihood based implicit generative model learning method, which focuses on learning expressive base distributions. 

We know from probability theory that in an implicit generative model as defined in Equation \eqref{eq:implmodel}, the output probability density is related to the base distribution via the cumulative density function: 
\begin{align}
	\pmodel[x][\theta, \phi] = \frac{\partial}{\partial x} \int_{ \{ x: f_\theta(h) \leq x \} } \pbase[p^0_\phi] dh, \label{eq:cdf}
\end{align}
where note that the base distribution is parametrized by $\phi$. The integral in Equation \eqref{eq:cdf} is not tractable in general, however if we have an invertible mapping $f_\theta(h)$, we can obtain an analytical expression for the density function of the model using the following formula \cite{devroye1986}: 
\begin{align}
	\pmodel[x][\theta, \phi] = \pbase[p^0_\phi][f^{-1}_{\theta}(x)]  V_\theta(x) \label{eq:pdfmodel},
	%\left | \det  J_\theta(x) \right |, \label{eq:pdfmodel}
\end{align} 
where $V_\theta(x):=  \left | \det \frac{\partial f^{-1}_{\theta}(x) }{\partial x} \right | = \left | \det \frac{\partial f_{\theta}(h) }{\partial h} \right |^{-1}$, which measures the volume change due to the transformation. It is possible to construct exactly invertible mappings using typical neural network mappings such as matrix multiplications and convolutions. Constraining the forward mapping to be exactly invertible is restrictive however, mainly because invertibility only holds for transformations which do not change the dimensionality. In section \ref{sec:twostage} we describe an algorithm which maximizes the model likelihood for a general mappings for which we also have an approximate inverse. 

%In this paper we explore the cases where the mapping $f_\theta(x)$ is invertible in section \ref{ }. We also look at the cases where $f_\theta(h)$ is approximately invertible, and propose an algorithm which approximately maximizes the maximum likelihood objective. 

\subsection{Maximum Likelihood for Implicit Generative Models}
If we work with invertible forward mappings, the optimization problem for maximum likelihood in an implicit generative model is the following: 
\begin{align}
	&\max_{\theta, \phi} \sum_n \log \pmodel[x_n][\theta,\phi], \notag \\
	= & \max_{\theta, \phi} \sum_n \log \pbase[p^0_\phi][f^{-1}_{\theta}(x_n)] + \log V_\theta(x_n), \label{eq:MLimp}
\end{align}
where the first term can be interpreted as maximizing the likelihood of the mappings $f^{-1}_\theta(x)$ in the base distribution space, and the volume term $V_\theta(x)$  ensures that the distribution properly normalized. If we think about this objective from a sampling perspective, in order to the generate plausible samples, the maximum likelihood objective tries to match the samples from the base distribution with the observations mapped to the base distribution space $f^{-1}_\theta(x)$.  

Note that in GANs, only the forward mapping parameters $\theta$ is optimized, and the base distribution is fixed to be simple unimodal distribution. Optimizing both the forward mapping parameters $\theta$ and a multi-modal base distribution constitutes the main idea in our paper. We argue that mapping a multimodal dataset onto a unimodal base distribution is harder to achieve than fitting a multimodal distribution on $f^{-1}_\theta(x)$. We demonstrate this in Figure \ref{fig:toyexample}. Using an invertible linear mapping $f_\theta(h)=W h$, where $h \in \mathbb R^2$, and $W \in \mathbb R^{2 \times 2}$, we show that on a two dimensional mixture of Gaussians example that, if we do maximum likelihood on the objective in Equation \eqref{eq:MLimp}, we fail to map the observations to the samples drawn from a fixed isotropic base distribution. However, as shown in Figure (b) if we set the base distribution as a flexible distribution such as mixture of Gaussians, and learn its parameters $\phi$, we are able to learn a much more accurate distribution. We also show that if we train the same mapping using the standard GAN formulation, we get the mode collapse behavior, where only one of the Gaussians is captured in the learned distribution. 

% make the point that if we dont do dimensionality reduction things are pointless. 
We acknowledge that in the cases where the forward mapping has the same dimensionality in the domain and range spaces (such as the example in Figure \ref{fig:toyexample}), learning an implicit generative model by maximizing Equation \eqref{eq:MLimp} is pointless, because we could have very well just fitted a mixture model on the data. For this reason, in the next section we propose the two stage learning algorithm which allows the use of forward mappings which change the dimensionality.

\newcommand{\fw}{1}
\begin{figure*}[h!]
	\begin{subfigure}[b]{0.49\textwidth}
		\includegraphics[width=\fw\textwidth]{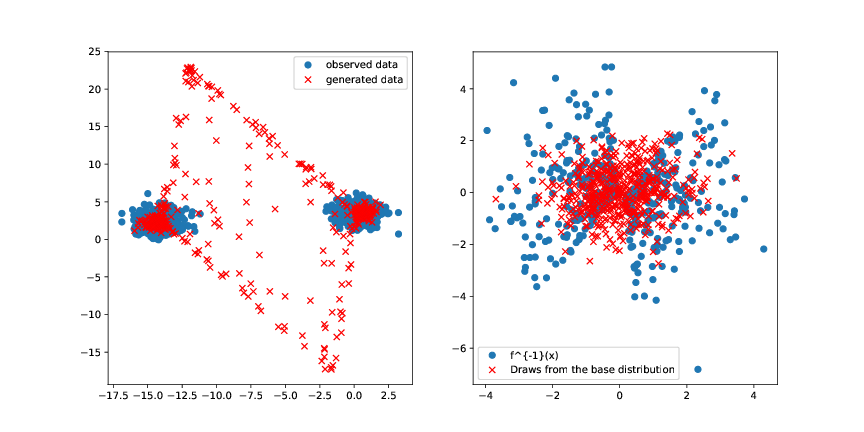}
		\caption{Using a simple and fixed base distribution}
	\end{subfigure}
	\begin{subfigure}[b]{0.49\textwidth}
		\includegraphics[width=\fw\textwidth]{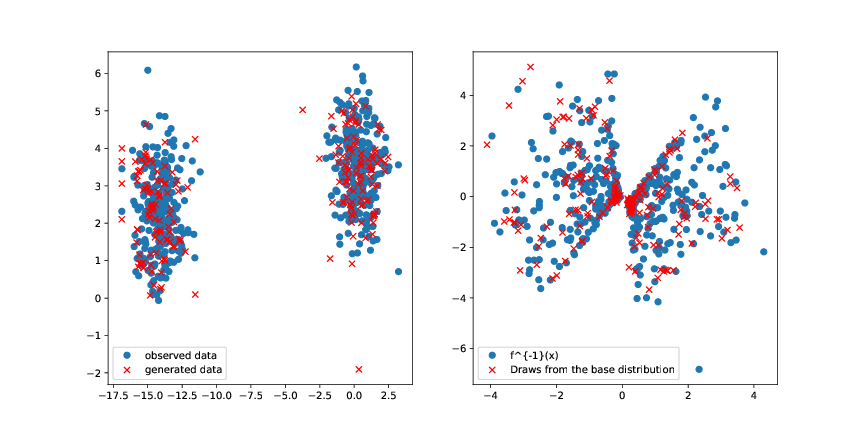}
		\caption{Learning the base distribution}
	\end{subfigure}
	\centering
	\begin{subfigure}[b]{0.49\textwidth}
		\includegraphics[width=\fw\textwidth]{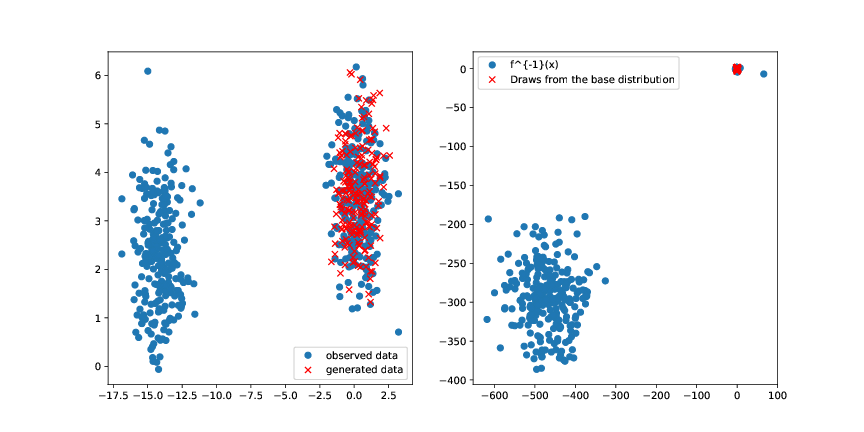}
		\caption{What GAN learns}
	\end{subfigure}
	\caption{We demonstrate the differences between our proposed method and two methods that use a simple base distributions on a two dimensional mixture of two Gaussians. (Maximizing \eqref{eq:MLimp} and GAN with a fixed isotropic Gaussian as base distribution).  In figure (a) we maximize the implicit model likelihood defined in \eqref{eq:MLimp} with respect to forward mapping parameters $\theta$. In figure (b) we fit a mixture of Gaussians for the same forward mapping as figure (a). In figure (c) we see what GAN learns for the same dataset. In Each figure, left plot shows the generated data overlaid on the observed dataset, and left plot show the samples from the base distribution overlaid on the observations mapped to the observation space $(f^{-1}_\theta(x))$.}
	\label{fig:toyexample}
\end{figure*}

\subsection{The Two Stage Algorithm} 
\label{sec:twostage}
In practice, we typically would like to have base distribution defined on a space which has lower dimensionality than the observation space. If this is the case, then it is impossible to have an exactly invertible mapping $f_\theta(h)$. It is however possible to have an approximately invertible forward mapping. This idea gives the hint for a very simple two stage maximum likelihood algorithm: We first fit an auto-encoder such that the error $\sum_n \| f_\theta(f^\text{enc}_\psi(x_n)) - x_n\|$ is minimized. Once the we are done with optimizing the autoencoder, we simply fit a base distribution on the embeddings $f^\text{enc}_\psi(x)$. The formal algorithm is specified in Algorithm \ref{algo:twostage}.

\begin{algorithm}[h!]
	\caption{The two stage implicit generative model learning algorithm}
	\label{algo:twostage}
	\begin{algorithmic}
		\STATE -Train the auto-encoder parameters $\theta$, $\psi$ such that:
			\begin{align*}
				\min_{\theta, \psi} \sum_n \| f_\theta(f^\text{enc}_\psi(x_n)) - x_n\| 
			\end{align*}
		\STATE -Fit the base distribution on the latent space such that: 
			\begin{align*}
				\max_{\phi} \sum_n \log \pbase[p^0_\phi][f^{enc}_{\psi}(x_n)] 
			\end{align*}
	\end{algorithmic}
\end{algorithm}

To see that this is a maximum likelihood algorithm, let us reconsider the likelihood function of the implicit generative model with the autoencoder: 
\begin{align}
	= & \max_{\phi} \sum_n \log \pbase[p^0_\phi][f^{enc}_{\psi}(x_n)] + \log V(x_n),\label{eq:MLimp}
	%\log \det \left | \tilde{J}_{\theta, \phi}(x_n) \right |, \label{eq:MLimp}
\end{align}
where we easily see that the base distribution parameters $\phi$ are independent of the volume term $V(x)$. Assuming that that the autoencoder learns a mapping close to the identity, we conclude that maximizing with respect to the base distribution parameters maximizes the model likelihood. 

\begin{figure*}[h!]
	\includegraphics[scale=0.5]{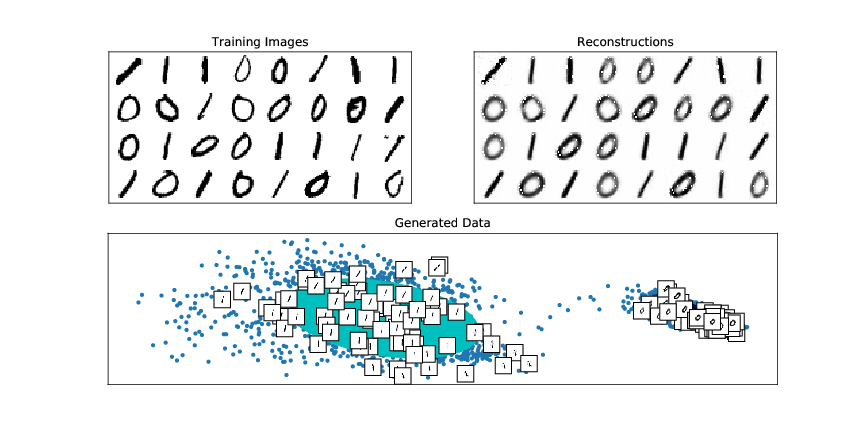}
	\caption{Demonstration of the two stage algorithm on a toy dataset with handwritten 0's and 1's. The purpose of this figure is to give a sense on how the proposed algorithm work. (Top row) Samples from the training set, and corresponding reconstructions. (Bottom row) Two dimensional embeddings of the training samples are shown with blue dots. We overlay sampled images. The solid color ellipses show the covariance components of the learned Gaussian mixture model for the base distribution. }
	\label{fig:mnisttoy}
\end{figure*}

Note that since the optimization for the forward mapping parameters $\theta$, and the base distribution is decoupled, it is easy to fit a multi-modal distribution for the base distribution on the embeddings $f^\text{enc}(x)$. One natural choice is to use a mixture distribution. We demonstrate this on handwritten zero and one digits from the MNIST dataset \cite{lecun-mnist} in Figure \ref{fig:mnisttoy}. We choose the dimensionality of the latent space $K=2$ to be able to visualize the base distribution space. We a three component Gaussian mixture model for this example. 

\subsection{Learning Generative Models for Sequential Data}
The framework we propose also offers the flexibility to learn distributions over sequences by simply learning a sequential distribution such as HMM on the latent representations. The likelihood of a sequence is expressed as follows: 
\begin{align*}
	\pmodel[x_{1:T}][\psi, \phi] = \prod_{t=1}^T p^0_\phi(f^\text{enc}_\psi(x_t) &| f^\text{enc}_\psi(x_{1:T-1})) V(x_t),  %\left | \det  J_\theta(x) \right |,
\end{align*}
where a sequence is denoted as $x_{1:T} := \{x_1, x_2, \dots x_T\}$. and thus $ f^\text{enc}_\psi(x_{1:T-1})= \{ f^\text{enc}_\psi(x_{1}),  f^\text{enc}_\psi(x_{2}), \dots, f^\text{enc}_\psi(x_{T-1})$\}. According to this density model, the observations $x_{1:T}$ are mapped to latent space independent from each other. This suggests that we can closely follow the two stage algorithm defined in Algorithm \ref{algo:twostage}: Same as before we first fit the autoencoder, and obtain the latent representations. In the second stage, instead of fitting an exchangeable model such as a mixture model, we fit a base distribution which models the temporal structure of the latent space. Potential options for such a distribution include Hidden Markov Models (HMMs), and RNNs, or convolutional models. In our audio experiments, we used HMMs with Gaussian emissions. 

%\subsection{Related Papers}
%I might use your help Sanmi here. 

% Ideas
% KL pq is broad - therefore we need an expressive model to be able not to produce junk
% show that we do max likelihood with a toy experiment - where try to learn jointly versus the two stage algorithm 
% show 2d mnist example maybe? (for base distribution learning concept)
% write that we cannot use this base distribution learning concept directly in VAEs or GANs - implicit distribution learning gives a natural framework
% you could show the 2d failing picture of VAEs

\section{Experiments}
% initial commit for section 2.3 
\subsection{Images}
We learn generative models on the MNIST \cite{lecun-mnist} (hand written digits) and CELEB-A \cite{celeba} (celebrity faces). We compare our algorithm (which we abbreviate with IML - Implicit Maximum Likelihood), with VAE, standard GAN and Wasserstein GAN. As the main quality metric, we compare likelihoods computed on a test set using kernel density estimator (KDE).  

%We apply the two stage learning algorithm in all datasets: We first fit an autoencoder to obtain the latent representations, and then we fit a Gaussian mixture model on the latent representation. 

For the MNIST dataset, we use an invertible perceptron in our approach to demonstrate that we can also use our approach to compute model likelihoods on the test set using the implicit generative model density function in Equation \eqref{eq:pdfmodel}. (Note that in general our framework allows non-invertible mappings: We use a general convolutional autoencoder for the CELEB-A dataset) The invertible perceptron we use for the MNIST dataset is defined as follows: 
\begin{align*}
	h_1 =& \tanh_\text{invt} \left ( \textbf{Linear}[K, 600](h) \right ), \\
	x =&  \sigma_\text{invt} \left ( \textbf{Linear}[600, 784](h_1) \right ),
\end{align*}
where $h \in \mathbb R^K$ denotes the latent representation, and $\textbf{Linear}[L_1, L_2](h)=W h + b$, $W\in \mathbb R^{L_2 \times L_1}$, $b \in \mathbb R^{L_2}$ represents a linear layer (we follow the pytorch API convention to denote the input and output dimensionalities). The invertible non-linearity functions are denoted with $\tanh^\text{invt}(.)$, and $\sigma^\text{invt}(.)$, which respectively stand for invertible tangent-hyperbolic and invertible sigmoid functions. We basically use the original non linearity in the invertible regime, and a linear function in the saturation regimes. Namely, for hyperbolic tangent we have the following function: 
\begin{align}
	\tanh_\text{invt}(t) =
	\begin{cases}
		c t - b & t \leq -1 + \epsilon  \\
		\tanh(t) & | t| \leq 1 - \epsilon \\
		c t + b & t \geq 1 - \epsilon 
	\end{cases}, 
\end{align}
We use $c=0.01$, and choose the bias term $b$, and the threshold $\epsilon$ so that the function is continuous and smooth (has a continuous first derivative). Similarly, the invertible sigmoid function is defined as follows: 
\begin{align}
	\sigma_\text{invt}(t) =
	\begin{cases}
		c t - b & t \leq \epsilon  \\
		\sigma(t) &  0 \leq t\leq 1 - \epsilon \\
		c t + b & t \geq 1 - \epsilon 
	\end{cases}, 
\end{align}
Note that it is straightforward to derive the inverse functions once the parameters of the non-linearities are set. Therefore the inverse network is defined as follows:
\begin{align*}
	h_1 =& \sigma_\text{invt}^{-1}( \textbf{Linear}^{-1}[784, 600](x)), \\ 
	h =& \tanh_\text{invt}^{-1}( \textbf{Linear}^{-1}[600, K](h_1)), \\ 
\end{align*}
where $\textbf{Linear}^{-1}[L_2, L_1](x) := (W^\top W)^{-1} W^\top (x - b)$. Note that the parameters $W$, $b$ are shared for a given forward and inverse Linear layers. To obtain the volume term due to the rectangular transformation, we note that the volume change due to the rectangular linear transformation in a linear layer is given by $\sqrt{\det (W^\top W})$ \cite{voldet}. Therefore to the correction term involves dividing the original pdf with this volume change (we note that the implicit model likelihood holds, because the mapping is approximately invertible due to the first step of the algorithm). 

%To obtain the Jacobian term, we consider the mapping $\mathcal{L}(W^\top x) = (W^\top W)^{-1} W^\top (x- b) : \mathbb R^K \to \mathbb R^K$, which maps $h$ to $W^\top x$. Then, 
%\begin{align*}
%	\left | \det \frac{\partial \mathcal{L}(x)}{\partial  (W^\top x) } \right | = \left | \det (W^\top W)^{-1} \right |	= \left | \det (W^\top W) \right |^{-1},
%\end{align*}
%where the last equality is due to the property of Jacobians of invertible functions. 
%To quantify the quality of the generated samples, in addition to computing the implicit model likelihood, we also compute Kernel density estimates. 

To do objective comparisons between models we compute Kernel density estimates (KDE) on the test set: For each batch, we sample 1000 points from the trained models, and represent the learned density as the sum of Kernel functions centered at these samples. We then compute the average score for all the test set. We use Gaussian Kernels, with bandwidth 0.01. Namely, the KDE scores we compute for the models are defined as follows: 
\begin{align*}
	&\text{KDE score} = \\
&\frac{1}{N_\text{test} N_\text{samples}} \sum_{n=1}^{N_\text{test}} \sum_{m=1}^{N_\text{samples}} \mathcal N(x^\text{test}_n; x^\text{sample}_m, 0.1I).  \notag 
\end{align*}
Notice that for small kernel bandwidth, the above objective is tantamount to computing the nearest neighbor distance for all test instances. To get high scores from this estimator, the observed samples need to capture the diversity of the test instances. Also note that this estimator is computing an estimate for $\text{KL}(p_\text{testset} \| p_\text{model})$, so this metric penalizes mode collapse. 

\begin{figure}[h!] 
	\includegraphics[trim=1cm 0cm 1cm 0 cm, clip, width=0.48\textwidth]{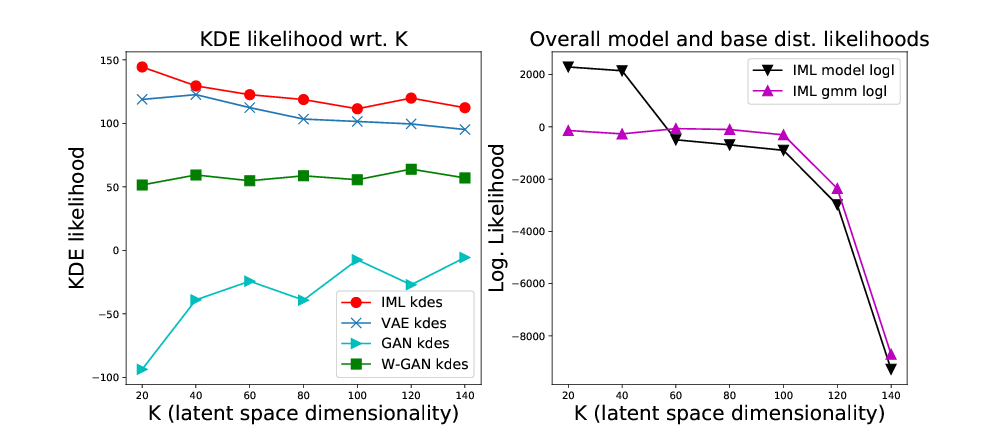}
	\caption{KDE likelihood with respect to the dimensionality of the latent space $K$ on the MNIST dataset. } 
	\label{fig:Kvskde}
\end{figure}

In the left panel of Figure \ref{fig:Kvskde}, we compare the KDE scores for our two-stage algorithm, GAN, Wasserstein GAN and VAE on the MNIST dataset. We use the standard training-test split defined in the pytorch data utilities (60000 training instances and 6000 test instances). We try 7 different latent dimensionality $K$ for all algorithms ranging from 20 to 140 with increments of 20. In our algorithm, we use a GMM with 30 full-covariance components for all $K$ values. We see that performance drops with increasing $K$, however we manage to stay better than VAEs and GANs. The performance drop is expected to happen with increasing $K$, because the density estimation problem in the latent space gets more difficult with increasing latent dimensionality. We would like to note that it possible to use a more complicated base distribution and compensate. 

In the right panel of Figure \ref{fig:Kvskde}, we compare the model likelihood computed with the implicit likelihood equation in \eqref{eq:pdfmodel} with the base distribution likelihood (the complete likelihood minus the Jacobian term). The purpose of this is to examine if there is a correlation between these quantities. As we pointed out before, our algorithm does not require an exactly invertible mapping, and as can be seen from the figure the base distribution likelihood is somewhat correlated with the overall model likelihood, and therefore can potentially be used as a proxy for the complete likelihood for mappings for which we don't know how to compute the Jacobian term. 

In Figure \ref{fig:nnsamples}, we show the random nearest neighbor samples for randomly selected test instances for all four algorithms in the top panel. We see that IML method is able to capture the diversity of the test instances well. On top of that we see much more definition in the generated images thanks to the multi-modal base distribution that we are using. As we earlier illustrated in Figure \ref{fig:toyexample}, using a simplistic base distribution causes a mismatch between the mappings to the latent space and the draws from the base distribution. Due to the simplistic distributions used in VAEs, and GANs we see that these approaches tend to generate more samples which do not resemble handwritten digits. We also observe that quality of the samples (and nearest neighbor samples) are correlated with the KDE metric.

\begin{figure*}
	\centering
	\includegraphics[scale=0.35]{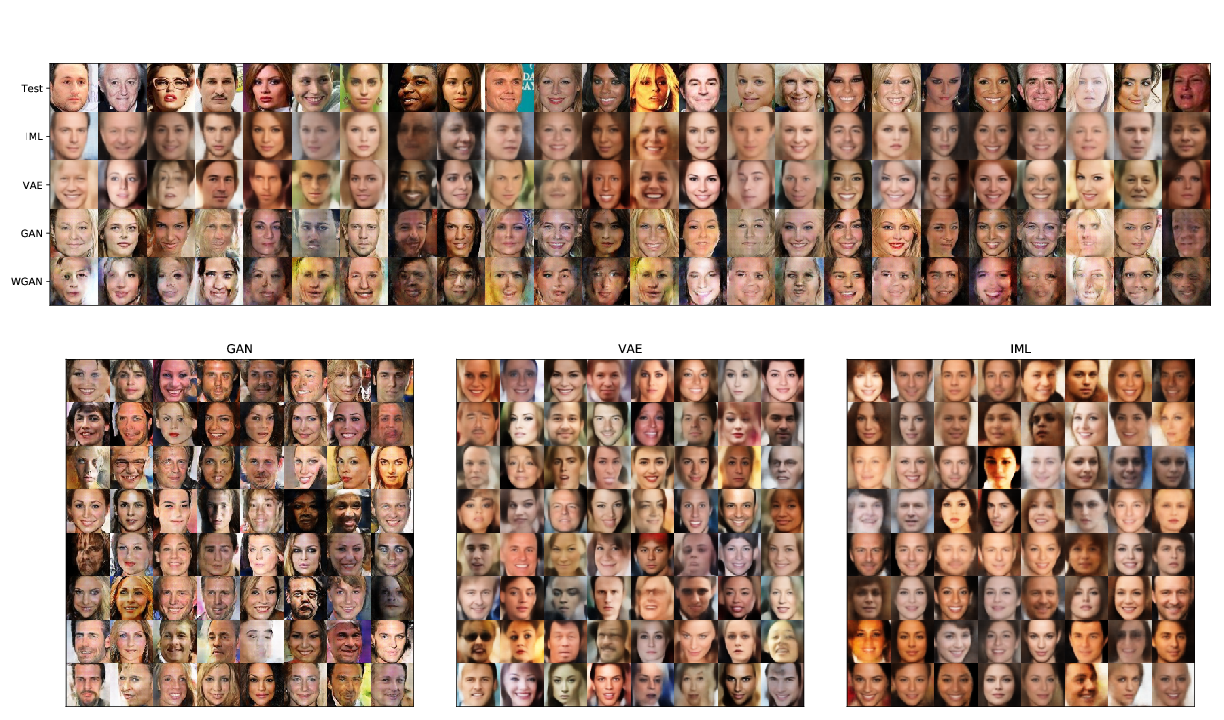}
	\caption{Samples from the CELEB-A dataset. (\textbf{top}) Generated nearest neighbor samples (nearest to test instances which are shown above) for four different algorithms. (\textbf{bottom-left}) Random images generated with GAN, (\textbf{bottom-middle}) Random images generated with VAE (\textbf{bottom-right}) Random Samples with IML, samples from the same cluster are grouped together.}
	\label{fig:nnsamples-celeba}
\end{figure*}

In Figure \ref{fig:nnsamples-celeba}, we do the same nearest neighbor sample measurement on the CELEB-A dataset. We have set the latent dimensionality as 100 for all algorithms. We cropped the images using a face detector, and resized them to size $64\times64$ in RGB space. We used 146209 such images for training, and 10000 images for test. We see that the proposed IML algorithm has more accurate nearest neighbor samples. We see that although the VAE is able to generate less distorted samples than GAN and WGAN, it's generated images contain much more distortion than IML, potentially because of the simplistic latent representation. The generated samples from IML contain much less distortion than GANs. 

For all algorithms we used the Adam optimizer \cite{Kingma2014}. As mentioned before, in the MNIST experiment, for IML we used the invertible network we introduced in this section. For GANs and VAE we used a standard one hidden layer perceptron with exact same sizes. Namely, the decoders maps $K$ dimensions into 600, and 600 dimensions then gets mapped into 784 dimensions (MNIST images are of size $28 \times 28$). We use the mirror image encoder for the VAE, that is we map 784 dimensions to 600, and that gets mapped into $K$ dimensional vectors for the mean and variance of the posterior. For the CELEB-A dataset, we used a 5 layer convolutional encoders and decoders (We used the basic DC-GAN \cite{Radford2015} generator architecture for all algorithms, with exact same parameter setting - only with the exception that for VAE the latent representations are obtained without passing through ReLU in order not to allow negative values as we use isotropic Gaussian as the prior). For W-GAN would like to point that we used to code published by the authors with the default parameter set-up. For GAN and VAE our code is based on code provided for pytorch examples. 
% then talk what you do with lr, networks, data split,  

\begin{table}[h]
\caption{Best KDE scores on test set for MNIST and CELEB-A datasets using 4 different algorithms. (Larger is better)}
\label{sample-table}
\begin{center}
\begin{tabular}{lll}
\multicolumn{1}{c}{\bf Algorithm}  &\multicolumn{1}{c}{\bf MNIST} &\multicolumn{1}{c}{\bf CELEB-A}\\
\hline \\
IML & 143 & -8318 \\
VAE & 132 & -11003\\
GAN & -5 & -11970 \\
WGAN & 64 & -12986\\
\end{tabular}
\end{center}
\end{table}

\newcommand{\scl}{0.35}
\begin{figure*}
	\centering
	\includegraphics[scale=\scl]{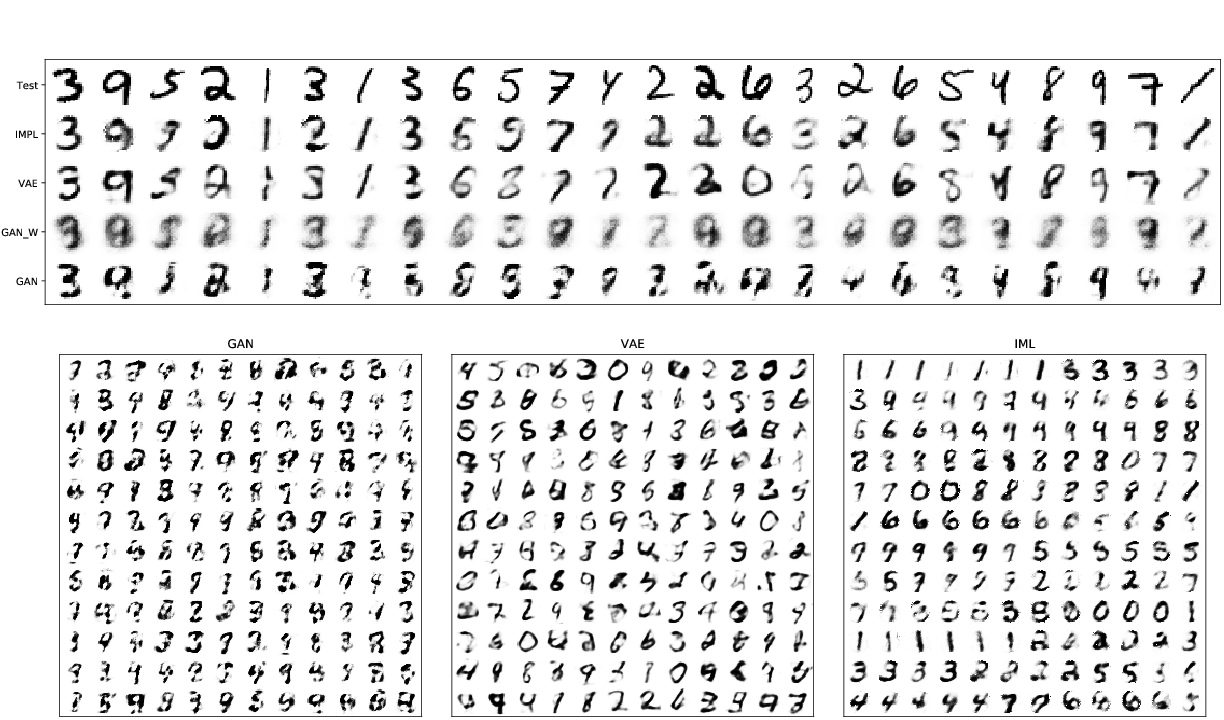}
	\caption{Samples from the MNIST dataset. (\textbf{top}) Generated nearest neighbor samples (nearest to test instances which are shown on the top row) for four different algorithms. (\textbf{bottom-left}) Random images generated with a GAN, (\textbf{bottom-middle}) Random images generated with a VAE, (\textbf{bottom-right}) Generated Samples with IML, samples from the same cluster are grouped together.}
	\label{fig:nnsamples}
\end{figure*}

\subsection{Audio}
To show that our algorithm can be used to learn a generative model for sequential data, we experiment with generating speech and music in the waveform domain. In all datasets, we work with audio with 8kHz sampling rate. We dissect the audio into 100ms long chunks, where consecutive chunks overlap by 50ms, and each window is multiplied by a Hann window. The autoencoder learns 80 dimensional latent representations for each chunk which is 800 samples long. We use three layer convolutional networks both in the encoder and decoder, where we use filters of length 200 samples. 

We fit an HMM to the extracted latent representations. We use 300 HMM states, where each state has a diagonal covariance Gaussian emission model. The random samples are obtained by sampling from the fitted HMMs, and passing the sampled latent representation through the decoder. To reconstruct the generated chunks as an audio waveform, we follow the overlap-add procedure \cite{Oppenheim2009}: We overlap the each generated by chunk by 50 percent and add. 

As a speech experiment, we learn a generative model over digit utterances. We work with the free spoken digit dataset \cite{spokendigits}. As the training sequence, we give the model a concatenated set of digit utterances. We consider the cases where the training data only contains one digit type, and the case where the training data contains all digits.
In Figure \ref{fig:spokendigits}, we show the spectrograms of generated digit utterances (this example contained all 10 digit types - we used 1000 utterances for training) along with spectrograms of the training digit utterances. Note that the generated digit utterances are generated in sequences (We generate one long sequence which contains multiple digits).  In Appendix, on figure \ref{fig:spokendigitsiso} we show three cases for the one-digit only training task. We see that we are able to learn a generative model over one digit with a some variety.

As the music experiment, we train a model on a 2 minute long violin piece. We downloaded the audio file for the violin etude in \url{https://www.youtube.com/watch?v=OuSI6t54KWY}. We show the spectrogram of the first 10 seconds of the piece and our generated sequence in Figure \ref{fig:violin}. We see from the spectrogram that the model is able to learn some musical structure, although there is additional background artifacts. The generated samples for the spoken digit utterances and the generated music sequence can be downloaded and listened from the following anonymous link: 
\url{https://www.dropbox.com/sh/6mvzf9ca1wl3uej/AAAkBTdNBumU61_mnMu7epDla?dl=0} (we suggest copy and pasting the link, and watching for spaces, also we suggest opening the files with vlc player if your native player does not work)

\newcommand{\wif}{0.45}
\begin{figure}[h!]
	\includegraphics[width=\wif\textwidth]{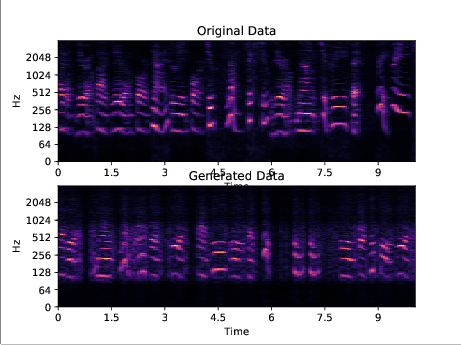}
	\caption{We illustrate the spectrograms for generated digits. Top figure contains the spectrogram for the true digit utterances, and below figure contains the spectrogram of the generated utterance.} 
	\label{fig:spokendigits}
\end{figure}

\begin{figure}[h!]
	\includegraphics[width=\wif\textwidth]{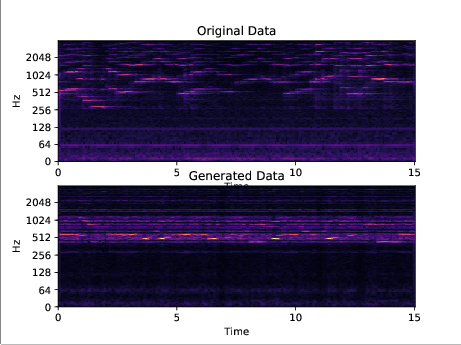}
	\caption{Excerpt from the spectrogram of the generated sequence learned from the violin etude}
	\label{fig:violin}
\end{figure}

\newcommand{\wiff}{0.40}

% comparisons of gan and our method both with invetible and non-invertible (conv.) networks ? 

\section{Discussion}
The algorithm we propose in this paper is very simple and effective. It is also principled in the sense that it performs maximum likelihood. We would like to emphasize that, compared the GANs the performance is much less sensitive to the network design choices and training parameters such as the learning rate. In author's experience, GANs are extremely sensitive to training parameters such as the learning rate. We have observed that decoupling the training of the base distribution from the neural network mapping makes the training much easier: In our approach it suffices to pick a small enough learning rate so that the encoder converges, and successfully embeds the data in a lower dimensional space. 

In our experience, VAE's seem to be easier to train (much less susceptible to hyperparameter choices). However, as we have seen in the results and figures, the simplistic choice for the base distribution results in distorted outputs. In our experiments we have used relatively more standard models to model the latent distribution, but it is possible to use complex methods such as Dirichlet Process Mixture models to obtain complicated base distributions. 

\bibliography{refs17}
\bibliographystyle{plain}

\newpage
\section{Appendix}
\subsection{Spectrograms of Individual Digits Utterances}
\begin{figure*}[h!]
	\includegraphics[width=\wif\textwidth]{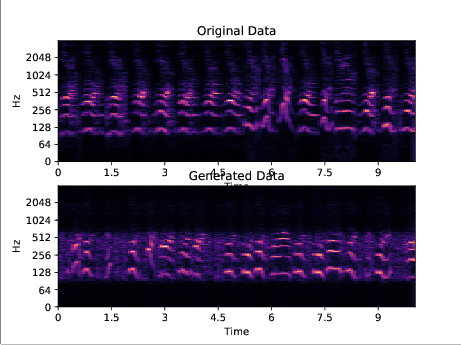}
	\includegraphics[width=\wif\textwidth]{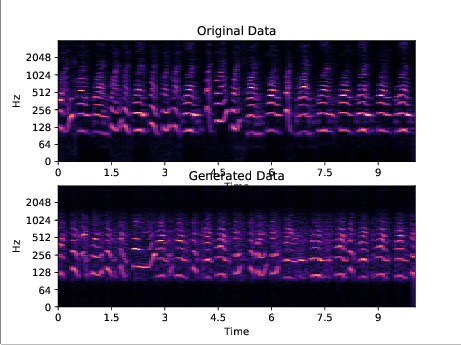}
	\centering
	\includegraphics[width=\wif\textwidth]{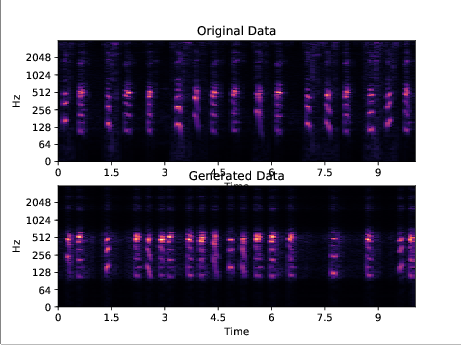}
	\caption{We illustrate the spectrograms for generated digits. \textbf{(top left)} Generated sequence for digit 0\textbf{(top right)} Generated sequence for digit 1 \textbf{(bottom)} Generated sequence for digit 6}
	\label{fig:spokendigitsiso}
\end{figure*}

Spectrograms fort the single type digit utterances are shown in Figure \ref{fig:spokendigitsiso}. 

\subsection{More Samples from CELEB-A} 
\begin{figure*}[ht]
	\includegraphics[scale=1]{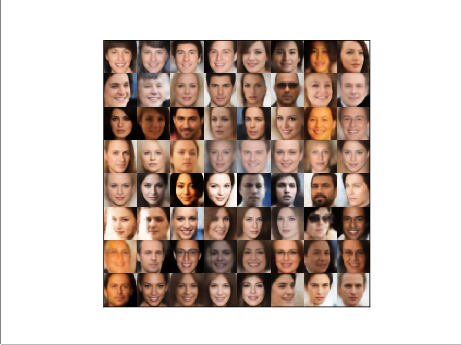}
	\caption{More Random Samples with IML on CELEB-A dataset.}
	\label{fig:imlceleba}
\end{figure*}

\begin{figure*}[ht]
	\includegraphics[scale=1]{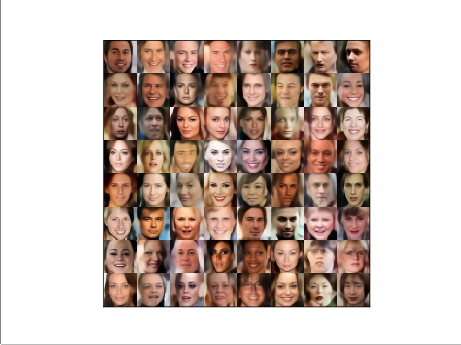}
	\caption{More Random Samples with VAE on CELEB-A dataset.}
	\label{fig:vaeceleba}
\end{figure*}

\begin{figure*}[h!]
	\includegraphics[scale=1]{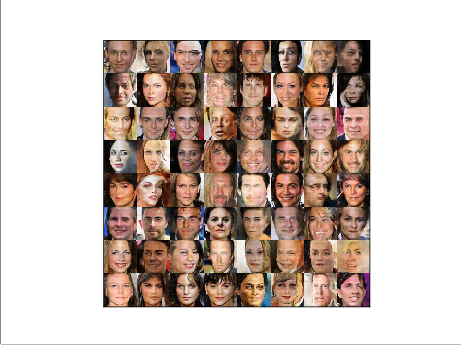}
	\caption{More Random Samples with GAN on CELEB-A dataset.}
	\label{fig:ganceleba}
\end{figure*}

\begin{figure*}[h!]
	\includegraphics[scale=1]{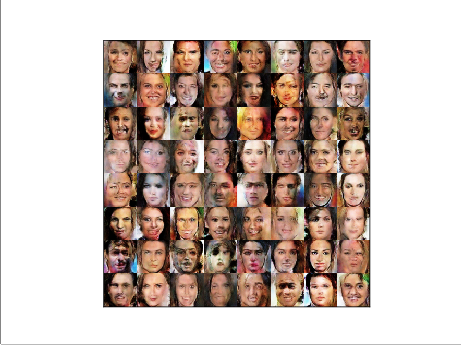}
	\caption{More Random Samples with Wasserstein-GAN on CELEB-A dataset.}
	\label{fig:wganceleba}
\end{figure*}

We show more random samples in Figures \ref{fig:imlceleba}, \ref{fig:vaeceleba}, \ref{fig:ganceleba}, \ref{fig:wganceleba} respectively with IML, VAE, GAN, and Wasserstein-GAN. 

\end{document}